\documentclass[letterpaper, 10 pt, conference]{ieeeconf}                                                            

\usepackage{CJKutf8}

\IEEEoverridecommandlockouts   

\usepackage{graphicx}
\usepackage{amsmath}

\begin{document}

\title{Vehicle Top Tag Assisted Vehicle-Road Cooperative Localization For Autonomous Public Buses}

\author{Hao Li$^*$, Bo Liu, Linbin Wang
\thanks{Hao Li, Linbin Wang are with the company \textit{Qingfei.AI}, Shanghai, 200240, and Hangzhou, 311121, China.}
\thanks{Bo Liu is with the Harbin Cambridge University, Harbin, 150069, China.}
\thanks{Corresponding author$^*$ : Hao Li (the CTO of the company \textit{Qingfei.AI}; e-mail: haoli@sjtu.edu.cn)}
}

\maketitle

\begin{abstract}
Accurate vehicle localization is indispensable to autonomous vehicles, but is difficult to realize in complicated application scenarios. Intersection scenarios that suffer from environmental shielding and crowded dynamic objects are especially crucial and challenging. To handle difficult intersection scenarios, the methodology of \textit{vehicle top tag assisted vehicle-road cooperative localization} or for short \textit{vehicle top tag assisted localization} is proposed. The proposed methodology has merits of satisfying all the \textit{feasibility}, \textit{reliability}, \textit{explainability}, \textit{society} and \textit{economy} concerns. Concrete solutions of vehicle top tag detection and vehicle top tag localization that instantiate the core part of the proposed methodology are presented. Simulation results are provided to demonstrate effectiveness of the presented solutions. The proposed methodology of vehicle top tag assisted localization also has the potential to be extended to a much wider range of practical applications than our intended ones involving autonomous public buses.
\end{abstract}

\IEEEpeerreviewmaketitle

\section{Introduction}

Accurate (i.e. centimetre-level) vehicle localization or for short vehicle localization is indispensable to autonomous vehicles. State-of-the-art (SOTA) vehicle localization systems normally rely on certain \textit{exteroceptive sensors} such as GNSS, LiDAR, and vision system (or camera), augmented by \textit{proprioceptive sensors} such as IMU. Relevant methods can be mainly categorized into GNSS based ones, LiDAR based ones, and vision based ones. These categories of vehicle localization methods are not mutually exclusive. In fact, merits of multi-sensor fusion architectures have long since been advocated \cite{Li2010itsc} \cite{Li2013TITS} \cite{Wan2018BaiduApollo} \cite{BaiX2025}. One may refer to \cite{Eskandarian2012} for a knowledge of vehicle localization methods in early years and refer to \cite{ShanX2023} for a survey of relevant methods in recent years.

\begin{figure}[h!]
\begin{center}
\includegraphics[width=0.97\columnwidth]{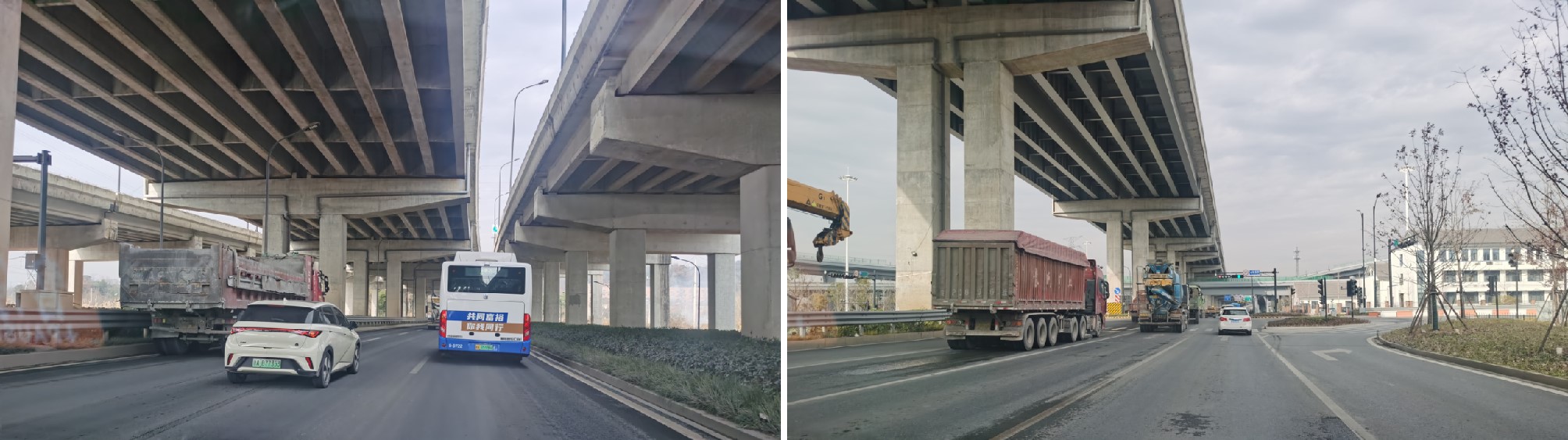}
\end{center}
\caption{Complicated application scenarios}
\label{fig:difficult_scenarios}
\end{figure}

Research fruits flourish in literature, yet no existing method can satisfy all the \textit{feasibility}, \textit{reliability}, \textit{explainability}, \textit{society} and \textit{economy} concerns for our dedicated ``vehicle-road-cloud'' projects, the core part of which consists in running \textit{autonomous public buses} in complicated application scenarios such as illustrated in Fig. \ref{fig:difficult_scenarios}. 

The application scenarios are especially challenging for two reasons: First, there is severe environmental shielding (against satellite signals), so high-accuracy GNSS based methods \cite{JingH2022} tend to fail. Second, there are many dynamic objects especially big vehicles such as trucks and engineering vehicles. Sometimes, the bus is even totally surrounded by big vehicles. LiDAR based methods \cite{YinH2024} tend to have unguaranteed performance in our application context. Similarly, vision based methods \cite{ChenW2022} \cite{ChenC2024} also suffer from difficulties due to severe occlusion caused by dynamic objects.

We by no means intend to exaggerate seemingly invalidity of existing methods. In fact, they are still somewhat useful to intended projects concerning autonomous public buses. For example, despite environmental shielding, GNSS based methods tend to enable rough global positioning as well. For the road part between neighbouring intersections, lane detection methods \cite{TangJ2021} \cite{QinZ2024} at least enable the autonomous public bus to follow the lane to navigate from one intersection to another, be there dynamic objects or not in the environment.

The indeed crucial parts of the complicated application scenarios are \textit{intersections}. They can be especially risky because they can be crowded by dynamic objects coming from and going in various directions. Besides, there is not any lane mark at intersections. When the autonomous public bus passes an intersection with environmental shielding and encounters view occlusion caused by big surrounding vehicles, it tends to get lost and fail to navigate desirably
\footnote{A naive thinking is that the vehicle may shortly rely on proprioceptive sensors such as IMU that are known to have moderate accumulated errors for short distances. Proprioceptive sensors are indeed so, but in terms of temporal relative localization \cite{Li2022RAL}. If the estimated vehicle orientation is not ideally aligned with the ground truth when the vehicle enters the intersection, which is very likely to happen in practice, vehicle localization based purely on proprioceptive sensors tends to still have unacceptable performance.}.

To handle difficult intersection scenarios, a technically sound idea is to add special road marks \cite{ZhuG2016} \cite{WangY2022} \cite{YangY2025} at intersections for guidance. But such kind of methods have difficulty in satisfying society concerns. First, road marks are strictly controlled by the government's road administration bureau. Anyone and even the bureau itself have no right to arbitrarily add any road mark especially alien road mark
\footnote{Special road marks or tags usually look alien or strange to humans.}. 
Second, alien road marks may cause confusion to human drivers and cause traffic disorder. Third, alien road marks may give people a feeling of visual intrusion and are socio-psychologically undesirable. The last two reasons somehow account for the first reason namely why road marks in traffic environment should be strictly controlled by the government.

Another technically sound idea is to take advantage of the V2X strategy which has been finding its way in more and more ``vehicle-road-cloud'' projects in recent years. Multi-vehicle cooperative localization \cite{Li2012ICVES} \cite{Li2013ITSMag} \cite{Li2014TITS} \cite{Li2024TITS} can be valuable to handle difficult intersection scenarios. However, according to our own experiences, difficulty of putting this into practice lies rather at policy and administration levels than at the technique level. Simply speaking, why should other vehicles cooperate with the autonomous public bus? After all, the government has no right to force social vehicles to do so.

Also as instantiation of the V2X strategy, vehicle-road cooperative localization \cite{HanY2023} \cite{GaoL2024} would be more sound
\footnote{When talking about V2X cooperative localization, we exclude wireless communication signal based methods \cite{HuangZ2024} which are usually researched in the communication domain, because even their SOTA performance is meter-level and far away from being satisfactory to autonomous public buses.}.
Thanks to ``vehicle-road-cloud'' projects, intersections are already equipped with perceptive roadside units (RSUs) especially visual RSUs. Autonomous public buses can fairly take advantage of perceptive RSUs at intersections to detect and indirectly localize the buses themselves. For the purpose, a key problem is how perceptive RSUs can accurately detect and localize buses. The problem may be generalized to the typical vehicle detection problem \cite{WangZ2023} \cite{LiangL2024} in the computer vision domain. However, even SOTA methods of general vehicle detection cannot claim to be a reliable solution with guaranteed localization accuracy in complicated application scenarios --- see Fig. \ref{fig:apriltag_detection}-right as example, which is especially difficult to general vehicle detection --- Despite lack of a desirable solution of general vehicle detection, we had better bear in mind that vehicle detection itself is not an ultimate objective for intended projects. It does not matter whether general vehicle detection or special kind of vehicle detection is used, only if autonomous public buses do can benefit from vehicle-road cooperative localization. We can fairly do something to autonomous public buses to facilitate detection and localization of them by perceptive RSUs. 

Following above reflection, we propose a methodology coined as \textit{vehicle top tag assisted vehicle-road cooperative localization} or for short \textit{vehicle top tag assisted localization}, clarification of which and whose merits will be postponed to Section \ref{sec:methodology}. Key technique points of the proposed methodology will be presented in more details in Section \ref{sec:tag_detection} and Section \ref{sec:tag_pose_estimation}, with practical applications oriented simulation results demonstrated and analysed in Section \ref{sec:simulation}, followed by a conclusion in Section \ref{sec:conclusion}.

\section{Methodology of vehicle top tag assisted localization}  \label{sec:methodology}

The primary contribution of this paper consists in the proposed methodology of vehicle top tag assisted localization, which is as follows: Install certain special tag plates on the vehicle top, an example of which is illustrated in Fig. \ref{fig:cam_four_views} and some other examples are illustrated in Fig. \ref{fig:vehicle_top_tags} in Appendix as well. As will be further explained in Section \ref{sec:tag_detection}, any other special tag can be designed and used, only if the special tag facilitates detection of it by visual RSUs and extraction of tag control points for vehicle top tag localization at intersections.

\begin{figure}[h!]
\begin{center}
\includegraphics[width=0.97\columnwidth]{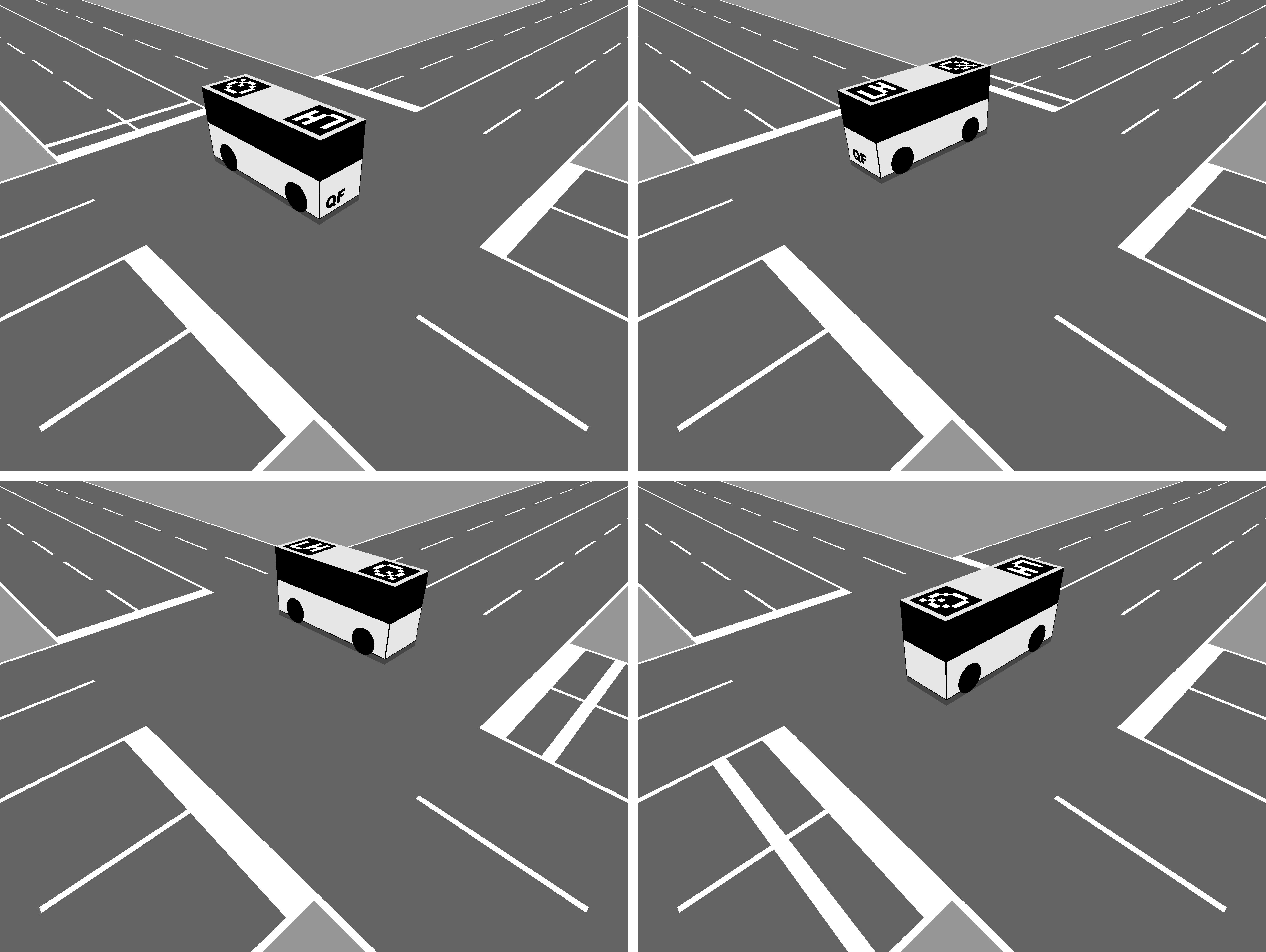}
\end{center}
\caption{Autonomous public bus with special vehicle top tags viewed from various perspectives by visual RSUs at the intersection}
\label{fig:cam_four_views}
\end{figure}

\begin{figure}[h!]
\begin{center}
\includegraphics[width=0.9\columnwidth]{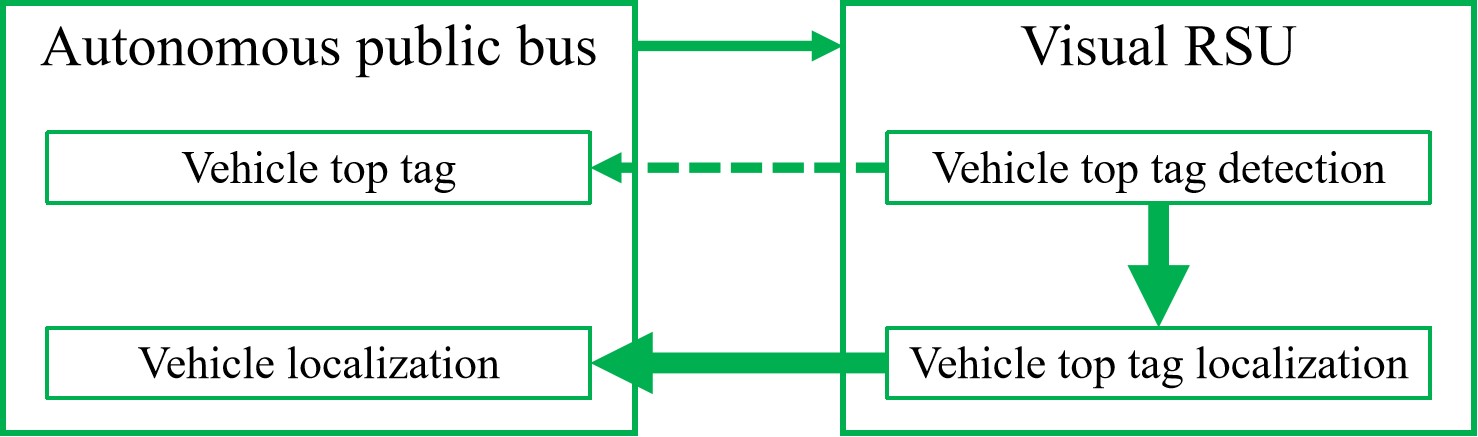}
\end{center}
\caption{Vehicle top tag assisted vehicle-road cooperative localization}
\label{fig:vehicle_road_cooperative_localization}
\end{figure}

For vehicle top tag assisted localization, when the autonomous public bus approaches an intersection, it communicates with visual RSUs at the intersection and inform them to initiate the process of vehicle top tag detection and localization. Upon initiating the process, visual RSUs try to detect vehicle top tags first, then perform vehicle top tag localization (i.e. vehicle top tag pose estimation) according to detected results, and next share pose estimates with the autonomous public bus. So the autonomous public bus can finally localize itself thanks to pose estimates shared by visual RSUs. When the autonomous public bus leaves the intersection, it informs visual RSUs there to close the process of vehicle top tag detection and localization (otherwise, visual RSUs will perform the process in vain and waste computational consumption). The working mechanism of the proposed methodology of vehicle top tag assisted localization is demonstrated in Fig. \ref{fig:vehicle_road_cooperative_localization}.

The proposed methodology of vehicle top tag assisted localization has following merits, corresponding to how it can satisfy all the necessary concerns namely the \textit{feasibility}, \textit{reliability}, \textit{explainability}, \textit{society} and \textit{economy} concerns.
\begin{itemize}
\item \textit{Feasibility} concern: Special tag plates that facilitate visual detection and localization are available and can be conveniently installed on bus tops. Besides, visual RSUs are used more and more at intersections to form intelligent intersections, which are important part of ``vehicle-road-cloud'' projects that have been developing rapidly (especially in China) in recent years. It has been more and more realized that vehicles, roads, and clouds inherently have mutually beneficial relationships and they should be developed holistically in the long run.
\item \textit{Reliability} concern: Special tags are so distinct and unique in the environment perceived by visual RSUs that normally neither false negative (missed detection) nor false positive (wrong detection) has any chance to exist. Besides, buses are among big vehicles in traffic environment and their high tops are unlikely to be occluded from visual RSUs which are usually installed at high places and perceive slanted-downward. 
\item \textit{explainability} concern: Special tags are highly structured and geometrically regular, so vehicle top tag detection and localization, as core part of the proposed methodology, are completely based on rigorous logic rules and geometry principles. In other words, they are completely explainable, which further accounts for reliability of the proposed methodology.
\item \textit{Society} concern: Bus tops are high and are beyond normal views of people (including human drivers) in traffic environment. So vehicle top tags have no risk of causing any feeling of visual intrusion to people in traffic environment and hence are socially desirable.
\item \textit{Economy} concern: As just mentioned above, ``vehicle-road-cloud'' projects have been developing rapidly and intelligent intersections equipped with visual RSUs are available. Here, we have no intention to argue why the ``vehicle-road-cloud'' holistic development mode is desirable for the whole society at macro-economy level. Instead, we would just like to point that in our application context at micro-economy level, the proposed methodology of vehicle top tag assisted localization enables autonomous public buses to effectively and reliably take advantage of available visual RSUs to handle difficult intersection scenarios, but only at an extra cost no other than the almost negligible cost of fabricating and installing vehicle top tags.
\end{itemize}

There are many aspects of engineering consideration for ``vehicle-road-cloud'' projects concerning autonomous public buses and how a complete instantiation of the proposed methodology is actually realized. In Section \ref{sec:tag_detection} and Section \ref{sec:tag_pose_estimation} that follow, we focus on clarifying concrete methods of vehicle top tag detection and vehicle top tag localization namely vehicle top tag pose estimation that instantiate the core part of the proposed methodology.

\section{Vehicle top tag detection}  \label{sec:tag_detection}

Special tags that facilitate visual detection and localization are available. AprilTags \cite{Olson2012} \cite{WangJ2016} are a kind of proper choice. Details of AprilTag detection (including tag control point extraction) can be found in \cite{Li2025GFCV}. We review them briefly here, with Fig. \ref{fig:cam_four_views}-top-left as demonstration example.

\begin{figure}[h!]
\begin{center}
\includegraphics[width=0.85\columnwidth]{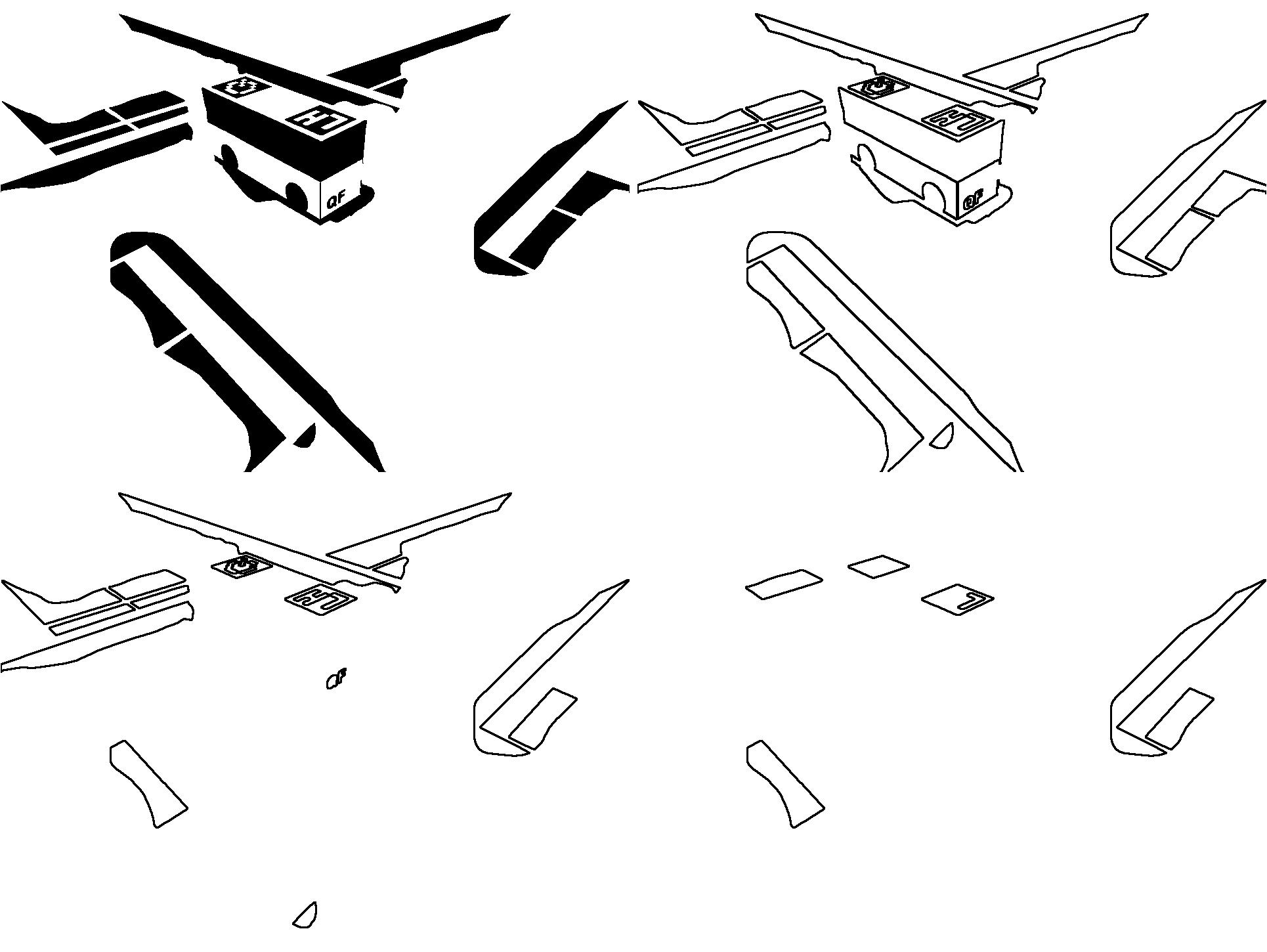}
\end{center}
\caption{Intermediate results: (top-left) binary image; (top-right) connected edges; (bottom-left) simple cycles; (bottom-right) quadrilaterals}
\label{fig:apriltag_one_detection_mid}
\end{figure}

\begin{itemize}
\item \textit{Adaptive thresholding}: Perform efficient adaptive thresholding of the image using the integral image technique and remove small foreground sets. The obtained binary image is demonstrated in Fig. \ref{fig:apriltag_one_detection_mid}-top-left.

\item \textit{Edge detection and connection}: Perform edge detection on the binary image and connect edge points into clusters, as demonstrated in Fig. \ref{fig:apriltag_one_detection_mid}-top-right.

\item \textit{Simple cycle extraction}: Check all edge clusters and only keep those that form simple cycles. Given a generic edge cluster, start from an edge point and move along the edge cluster in the direction such that the non-edge neighbour is always on the left. Verify whether the trip in the move-along-left way can traverse the edge points of the cluster exactly once and finally return to the starting edge point. If so, the edge cluster is extracted as a simple cycle and is kept as a quadrilateral candidate; otherwise, it is discarded. Extracted simple cycles are demonstrated in Fig. \ref{fig:apriltag_one_detection_mid}-bottom-left.

\item \textit{Quadrilateral verification}: For each simple cycle, verify if it is actually a quadrilateral, by checking whether it has four corners exactly. Given a generic edge point $\mathbf{P}$ of it, take two other edge points $\mathbf{P}_L$ and $\mathbf{P}_R$ on the two sides of $\mathbf{P}$ respectively and with certain distance away from $\mathbf{P}$. Compute the inner product of the vectors $\mathbf{P}_R - \mathbf{P}$ and $\mathbf{P} - \mathbf{P}_L$. Compute inner products corresponding to edge points of the simple cycle, apply non-minimum suppression to the inner products (a Gaussian filtering of the inner products before non-minimum suppression would help). If there are exactly four inner product valleys below certain threshold, then the simple cycle is verified as a quadrilateral and kept. Verified quadrilaterals are demonstrated in Fig. \ref{fig:apriltag_one_detection_mid}-bottom-right. To refine localization of quadrilateral corners, apply line fitting to each side of the quadrilateral and take intersections of the four fitted lines as the refined results.

\item \textit{AprilTag decoding and identification}: The AprilTag consists of a square matrix of cells, the boundary layer cells of which are all black cells surrounded by an even outer boundary layer of white or light-color background. To reveal the internal matrix of black-white cells (i.e. AprilTag code), compute the homography between the tag plane and the image plane with the extracted quadrilateral corners. Use the computed homography to project the internal matrix of black-white cells onto the image plane to obtain their corresponding image values (a sparse box filtering based on the already established integral image would help) and know the quadrilateral decoding result. 
Finally, match the decoding result with the bag of AprilTag codes
\footnote{When the autonomous public bus approaches the intersection, it shares the bag of its own tag codes with visual RSUs at the intersection.}
according to the Hamming distance. If the decoding result can be matched with one among the bag of AprilTag codes (considering the four potential rotations of $0^{\circ}$, $90^{\circ}$, $180^{\circ}$, $270^{\circ}$), then the quadrilateral is identified as the corresponding AprilTag and its four corners can be ordered correctly as well.
\end{itemize}

\begin{figure}[h!]
\begin{center}
\includegraphics[width=0.99\columnwidth]{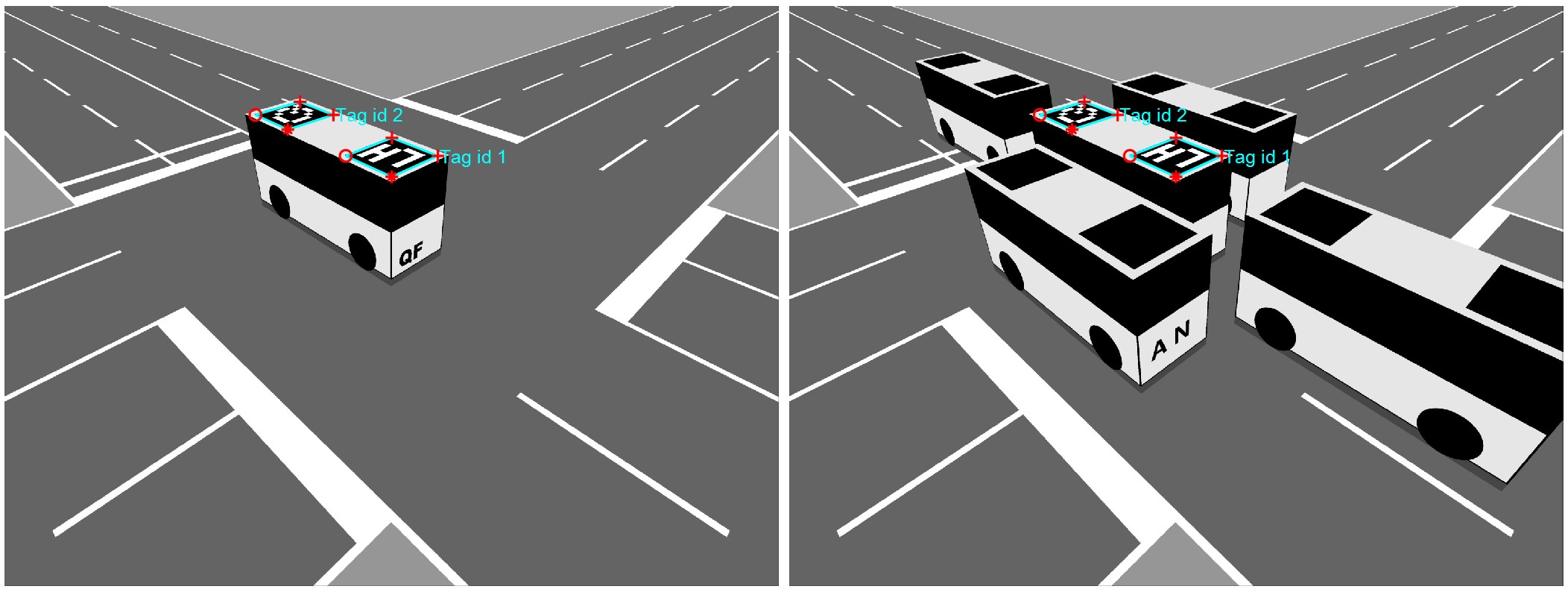}
\end{center}
\caption{AprilTag detection including tag control point extraction: (left) autonomous public bus itself; (right) surrounded by big vehicles}
\label{fig:apriltag_detection}
\end{figure}

Even with rather tolerant algorithm parameters for sake of guaranteeing no false negative, only few false positives usually could survive after the quadrilateral verification step. Since AprilTags are so distinct and unique in the perceived environment, the last step of AprilTag decoding and identification will not leave false positives any chance to exist. Effects of AprilTag detection are demonstrated in Fig. \ref{fig:apriltag_detection}. Even when the bus is surrounded and severely occluded by big vehicles, its top tags can still be effectively detected.

The reviewed AprilTag detection method only involves basic computational geometry techniques and a little 3D computer vision techniques. Its computational complexity is of order $O(\mbox{image size})$ which is already moderate. Besides, one can take advantage of pyramid-style hierarchical processing and visual tracking \textit{a priori} to even largely accelerate the detection. Relevant engineering details are omitted here.

Here, we have no intention to advocate the reviewed AprilTag detection method. For the proposed methodology of vehicle top tag assisted localization, any concrete algorithm of AprilTag detection does not matter. Above presentation is just to let readers know that AprilTag detection is realizable and give readers an idea of how it is realized. On the other hand, any appropriate method of AprilTag detection can be incorporated into the proposed methodology. In fact, even concrete choosing of vehicle top tags does not matter either. Any other kinds of special tags can be fairly adopted for the proposed methodology, only if they facilitate detection of them by visual RSUs and extraction of tag control points for vehicle top tag pose estimation.

No matter which concrete kind of vehicle top tags and relevant detection methods are used, the output is a set of extracted tag control points. Then given the tag control points, how to estimate the vehicle top tag pose? This is an even more important problem, solutions to which are given in Section \ref{sec:tag_pose_estimation}. 

\section{Vehicle top tag pose estimation}  \label{sec:tag_pose_estimation}

For an AprilTag, its control points are its four outer corners, as illustrated by the circle, plus, star marks namely ``$\circ$'', ``$+$'', ``$*$'' in Fig. \ref{fig:apriltag_detection}. The distinct marks are shown to convey that the order of tag control points can also be known via AprilTag detection. Positions of the four physical corners of the AprilTag in the vehicle top tag coordinates system
\footnote{A conventional way of establishing the vehicle coordinates system is to let the origin be located at the vehicle center projection on the ground plane and let the $x$-axis, $y$-axis, $z$-axis face forward, left-side, upward respectively. The vehicle top tag coordinates system differs from the conventional vehicle coordinates system only by a vehicle height, with its origin located at the vehicle center projection on the vehicle top tag plane instead of the ground plane. Since autonomous navigation depends on the vehicle horizontal pose and the two coordinates systems have no difference in their horizontal poses, for analysis convenience, we set the vehicle coordinates system identical to the vehicle top tag coordinates system. Whenever vehicle top tag pose estimation is mentioned throughout this paper, vehicle pose estimation (i.e. vehicle localization) is mentioned implicitly as well.}
can be known \textit{a priori}. For example, given a single AprilTag installed at the vehicle top center (see Fig. \ref{fig:vehicle_top_tags}-top-left) and the AprilTag width $2 w$, then the four physical corners always have the fixed vehicle top tag coordinates
\begin{align}  \label{eq:AprilTag_veh_top_xy}
&(x_1, y_1, z_1) = (-w, -w, 0), \quad (x_2, y_2, z_2) = (-w, w, 0),  \nonumber \\
&(x_3, y_3, z_3) = (w, w, 0), \quad (x_4, y_4, z_4) = (w, -w, 0).
\end{align}
Suppose their image coordinates extracted via AprilTag detection are respectively
\begin{equation}  \label{eq:AprilTag_img_uv}
(u_1, v_1), \quad (u_2, v_2), \quad (u_3, v_3), \quad (u_4, v_4).
\end{equation}
Suppose certain world coordinates system is established at the intersection. For a generic visual RSU, the intrinsic matrix $\mathbf{A}$ and the extrinsic parameters (i.e. the rotation matrices $\mathbf{R}_{cw}$, $\mathbf{R}_{wc}$ and the translation vectors $\mathbf{T}_{cw}$, $\mathbf{T}_{wc}$) between it and the world coordinates system can be calibrated \textit{a priori} and treated as knowns. We provide solutions to vehicle top tag pose estimation given above knowns. One can refer to the first author's book \cite{Li2025GFCV} for necessary geometry fundamentals that support this section \textit{theoretically}.

\subsection{Basic version of vehicle top tag pose estimation}

To facilitate understanding, readers may consider the single AprilTag configuration illustrated in Fig. \ref{fig:vehicle_top_tags}-top-left as example and simply treat $x_i$, $y_i$, $u_i$, $v_i$ ($i \in \{1 \cdots n\}$) in following formalisms as specified concretely in (\ref{eq:AprilTag_veh_top_xy}) and (\ref{eq:AprilTag_img_uv}) with $n = 4$, though the following formalisms are general and also applicable to multi-AprilTag configurations.

Compute the homography between the vehicle top tag plane and the image plane by solving the linear equation
\begin{equation}  \label{eq:homography_calibration_compact}
\mathbf{L} \mathbf{h}_r = \mathbf{0},
\end{equation}
where
\begin{align*}
\mathbf{L} \equiv \begin{bmatrix} x_{1} & y_{1} & 1 & 0 & 0 & 0 & -u_1 x_{1} & -u_1 y_{1} & -u_1 \\ 0 & 0 & 0 &  x_{1} & y_{1} & 1 & -v_1 x_{1} & -v_1 y_{1} & -v_1 \\ \vdots & \vdots & \vdots & \vdots & \vdots & \vdots & \vdots & \vdots & \vdots \\ x_{n} & y_{n} & 1 & 0 & 0 & 0 & -u_n x_{n} & -u_n y_{n} & -u_n \\ 0 & 0 & 0 &  x_{n} & y_{n} & 1 & -v_n x_{n} & -v_n y_{n} & -v_n \end{bmatrix}
\end{align*}
and
\begin{align*}
\mathbf{h}_r \equiv \begin{bmatrix} \mathbf{e}_1^\mathrm{T} \mathbf{H} & \mathbf{e}_2^\mathrm{T} \mathbf{H} & \mathbf{e}_3^\mathrm{T} \mathbf{H} \end{bmatrix}^\mathrm{T} 
\end{align*}
is the row-concatenated vector of the homography matrix $\mathbf{H}$. The solution of $\mathbf{h}_r$ is the normalized eigenvector associated with the smallest eigenvalue of $\mathbf{L}^\mathrm{T} \mathbf{L}$. Once $\mathbf{h}_r$ is solved, the homography matrix $\mathbf{H}$ can be recovered from $\mathbf{h}_r$.

Compute the relative pose between the vehicle top tag coordinates system and the camera (visual RSU) coordinates system, namely the rotation matrix $\mathbf{R}_{tc}$ and the translation vector $\mathbf{T}_{tc}$ from the former to the latter, the rotation matrix $\mathbf{R}_{ct}$ and the translation vector $\mathbf{T}_{ct}$ from the latter to the former, sequentially as follows
\begin{subequations}  \label{eq:basic_RTtc+RTct}
\begin{align}
\mathbf{R}_{tc} \mathbf{e}_1 :&= \frac{\mathbf{A}^{-1} \mathbf{H} \mathbf{e}_1}{\| \mathbf{A}^{-1} \mathbf{H} \mathbf{e}_1 \|_2},  \\
\mathbf{R}_{tc} \mathbf{e}_2 :&= \frac{\mathbf{A}^{-1} \mathbf{H} \mathbf{e}_2}{\| \mathbf{A}^{-1} \mathbf{H} \mathbf{e}_2 \|_2} ,  \\
\mathbf{R}_{tc} \mathbf{e}_3 :&= (\mathbf{R}_{tc} \mathbf{e}_1) \times (\mathbf{R}_{tc} \mathbf{e}_2),  \\
\mathbf{w}_{tc} &= \mathbf{R}^{-1}(\mathbf{R}_{tc}), \\
\mathbf{R}_{tc} :&= \mathbf{R}(\mathbf{w}_{tc}) \qquad (\mbox{update the rotation matrix}),  \\
\lambda &= \sqrt{\| \mathbf{A}^{-1} \mathbf{H} \mathbf{e}_1 \|_2 \cdot \| \mathbf{A}^{-1} \mathbf{H} \mathbf{e}_2 \|_2},  \\
\mathbf{T}_{tc} &= \frac{1}{\lambda} \mathbf{A}^{-1} \mathbf{H} \mathbf{e}_3,  \\
\mathbf{R}_{ct} &= \mathbf{R}_{tc}^\mathrm{T},  \\
\mathbf{T}_{ct} &= -\mathbf{R}_{tc}^\mathrm{T} \mathbf{T}_{tc},
\end{align}
\end{subequations}
where $:=$ denotes value assignment rather than equality, $\mathbf{R}(\cdot)$ denotes the rotation function that maps a rotation vector to its associated rotation matrix, and $\mathbf{R}^{-1}(\cdot)$ denotes the inverse rotation function that maps a (quasi) rotation matrix to its associated rotation vector. 

Finally, compute the pose of the vehicle top tag coordinates system with respect to the world coordinates system, namely the rotation matrix $\mathbf{R}_{tw}$ (or the rotation vector $\mathbf{w}_{tw}$) and the translation vector $\mathbf{T}_{tw}$ from the former to the latter
\begin{align}  \label{eq:basic_RTtw}
& \begin{bmatrix} \mathbf{R}_{tw} & \mathbf{T}_{tw} \\ \mathbf{0}^{\mathrm{T}} & 1 \end{bmatrix} = \begin{bmatrix} \mathbf{R}_{cw} & \mathbf{T}_{cw} \\ \mathbf{0}^{\mathrm{T}} & 1 \end{bmatrix} \begin{bmatrix} \mathbf{R}_{tc} & \mathbf{T}_{tc} \\ \mathbf{0}^{\mathrm{T}} & 1 \end{bmatrix}  \iff   \nonumber \\
& \mathbf{w}_{tw} = \mathbf{R}^{-1}(\mathbf{R}_{cw} \mathbf{R}_{tc}), \quad \mathbf{T}_{tw} = \mathbf{R}_{cw} \mathbf{T}_{tc} + \mathbf{T}_{cw}. 
\end{align}
Retrieve the horizontal part of the vehicle top tag pose as
\begin{equation}  \label{eq:basic_bus_pose}
\begin{bmatrix} x_w \\ y_w \\ \phi_w \end{bmatrix} = \begin{bmatrix} \mathbf{e}_1^\mathrm{T} \mathbf{T}_{tw} \\ \mathbf{e}_2^\mathrm{T} \mathbf{T}_{tw} \\ \mathbf{e}_3^\mathrm{T} \mathbf{w}_{tw} \end{bmatrix} = \begin{bmatrix} \mathbf{e}_1^\mathrm{T} \mathbf{T}_{tw} \\ \mathbf{e}_2^\mathrm{T} \mathbf{T}_{tw} \\ \mathbf{e}_3^\mathrm{T} \mathbf{R}^{-1}(\mathbf{R}_{tw}) \end{bmatrix}.
\end{equation}
As already explained in the previous footnote, the horizontal pose given in (\ref{eq:basic_bus_pose}) is right the vehicle horizontal pose (i.e. ``vehicle pose''). We refer to the computations (\ref{eq:homography_calibration_compact}), (\ref{eq:basic_RTtc+RTct}), (\ref{eq:basic_RTtw}), (\ref{eq:basic_bus_pose}) as the \textit{basic solution version}, simply denoted ``\textit{Bas.}''.

\subsection{Pose optimization considering hard geometry constraint}

As the bus height (denoted $h$) is fixed and can be known \textit{a priori}, there is a natural geometry constraint for the vehicle top tag plane, namely it is located on the plane
\begin{equation}  \label{eq:h_geometry_constraint}
z_w = h.
\end{equation}
In the light of the geometry constraint (\ref{eq:h_geometry_constraint}), positions of the AprilTag corners can also be computed as follows
\begin{align}  \label{eq:hard_opt_xy_initial}
\lambda \begin{bmatrix} u \\ v \\ 1 \end{bmatrix} &= \mathbf{A} \begin{bmatrix} \mathbf{R}_{wc} \mathbf{e}_1 & \mathbf{R}_{wc} \mathbf{e}_2 & \mathbf{R}_{wc} \mathbf{e}_3 & \mathbf{T}_{wc} \end{bmatrix} \begin{bmatrix} x_w \\ y_w \\ z_w \\ 1 \end{bmatrix} \implies  \nonumber \\
\begin{bmatrix} x_w \\ y_w \\ -\lambda \end{bmatrix} &= \begin{bmatrix} \mathbf{C}_1 & \mathbf{C}_2 & \begin{bmatrix} u \\ v \\ 1 \end{bmatrix} \end{bmatrix}^{-1} (- \mathbf{C}_3 h - \mathbf{A} \mathbf{T}_{wc}),
\end{align}
where
\begin{align*}
\mathbf{C}_1 \equiv \mathbf{A} \mathbf{R}_{wc} \mathbf{e}_1, \quad \mathbf{C}_2 \equiv \mathbf{A} \mathbf{R}_{wc} \mathbf{e}_2, \quad \mathbf{C}_3 \equiv \mathbf{A} \mathbf{R}_{wc} \mathbf{e}_3.
\end{align*}
Substitute $u_i$, $v_i$ ($i \in \{1 \cdots n\}$) respective for $u$, $v$ in (\ref{eq:hard_opt_xy_initial}) to obtain their corresponding world coordinates $x_{w,i}$, $y_{w,i}$, noting that $z_{w,i} = h$ is already known. 

Now we consider the hard geometry constraint that the vehicle top tag plane is constrained on the plane (\ref{eq:h_geometry_constraint}) strictly and latter we consider the soft geometry constraint. Under the hard geometry constraint, the vehicle top tag pose is actually its horizontal pose. Set the vehicle top tag initial pose as
\begin{equation}  \label{eq:hard_opt_initial_pose}
\begin{bmatrix} x_w^{init} \\ y_w^{init} \\ \phi_w^{init} \end{bmatrix} = \begin{bmatrix} \bar{x}_w \\ \bar{y}_w \\ \mathbf{e}_3^\mathrm{T} \mathbf{w}_{tw} \end{bmatrix} = \begin{bmatrix} (\sum_{i=1}^n x_{w,i})/n \\ (\sum_{i=1}^n y_{w,i})/n \\ \mathbf{e}_3^\mathrm{T} \mathbf{R}^{-1}(\mathbf{R}_{tw}) \end{bmatrix},
\end{equation}
where the initial orientation $\phi_w^{init}$ is still computed via (\ref{eq:basic_bus_pose}). Then perform optimization of the vehicle top tag pose as
\begin{equation}  \label{eq:hard_opt_pos}
\{x_w, y_w, \phi_w\} = \arg \min_{x_w, y_w, \phi_w} \sum_{i=1}^n \| \begin{bmatrix} u_i^H \\ v_i^H \end{bmatrix} - \begin{bmatrix} u_i \\ v_i \end{bmatrix} \|_2^2,
\end{equation}
where $\{u_i^H, v_i^H\}$ are functions in terms of $\{x_w, y_w, \phi_w\}$ that are computed sequentially as follows
\begin{subequations}  \label{eq:hard_opt_uv_proj}
\begin{align}
\mathbf{R}_{tw} &= \mathbf{R}(\begin{bmatrix} 0 & 0 & \phi_w \end{bmatrix}^\mathrm{T}),  \\
\mathbf{T}_{tw} &= \begin{bmatrix} x_w & y_w & h \end{bmatrix}^\mathrm{T},  \\
\lambda \begin{bmatrix} u_i^H \\ v_i^H \\ 1 \end{bmatrix} &= \mathbf{A} \begin{bmatrix} \mathbf{R}_{wc} & \mathbf{T}_{wc} \end{bmatrix} \begin{bmatrix} \mathbf{R}_{tw} & \mathbf{T}_{tw} \\ \mathbf{0}^\mathrm{T} & 1 \end{bmatrix} \begin{bmatrix} x_i \\ y_i \\ z_i \\ 1 \end{bmatrix}.
\end{align}
\end{subequations}
We refer to (\ref{eq:hard_opt_pos}) with needed pre-computations as the \textit{hard optimization solution version}, simply denoted ``\textit{H.Opt.}''.

\subsection{Pose optimization considering soft geometry constraint}

The the geometry constraint (\ref{eq:h_geometry_constraint}) may be slightly violated due to moderate disturbance (mainly up-down disturbance) as the autonomous public bus moves. Instead of (\ref{eq:hard_opt_pos}), perform optimization of the vehicle top tag pose $\mathbf{w T} \equiv \{\mathbf{w}_{tw}, \mathbf{T}_{tw}\}$ considering soft geometry constraint as
\begin{equation}  \label{eq:soft_opt_pos}
\mathbf{w T} = \arg \min_{\mathbf{w T}} \sum_{i=1}^n ( \| \begin{bmatrix} u_i^S \\ v_i^S \end{bmatrix} - \begin{bmatrix} u_i \\ v_i \end{bmatrix} \|_2^2 + \mu^2 |z_{w,i}^S - h|^2 ),
\end{equation}
the initial pose to which is augmented from that given in (\ref{eq:hard_opt_initial_pose})
\begin{align*}
\mathbf{w}_{tw}^{init} = \begin{bmatrix} 0 & 0 & \phi_w^{init} \end{bmatrix}^\mathrm{T}, \quad \mathbf{T}_{tw}^{init} = \begin{bmatrix} x_w^{init} & y_w^{init} & h \end{bmatrix}^\mathrm{T}.
\end{align*}
In (\ref{eq:soft_opt_pos}), $\{u_i^S, v_i^S\}$ are functions in terms of $\mathbf{w T}$ that are computed sequentially as follows
\begin{subequations}  \label{eq:soft_opt_uv_proj}
\begin{align}
\mathbf{R}_{tw} &= \mathbf{R}(\mathbf{w}_{tw}),  \\
\begin{bmatrix} x_{w,i}^S \\ y_{w,i}^S \\ z_{w,i}^S \end{bmatrix} &= \mathbf{R}_{tw} \begin{bmatrix} x_i \\ y_i \\ z_i \end{bmatrix} + \mathbf{T}_{tw},  \\
\lambda \begin{bmatrix} u_i^S \\ v_i^S \\ 1 \end{bmatrix} &= \mathbf{A} \begin{bmatrix} \mathbf{R}_{wc} & \mathbf{T}_{wc} \end{bmatrix} \begin{bmatrix} x_{w,i}^S \\ y_{w,i}^S \\ z_{w,i}^S \\ 1 \end{bmatrix}.
\end{align}
\end{subequations}
The parameter $\mu$ is to balance the image projection residuals and AprilTag corner height discrepancies. We simply fix $\mu$ and always set $\mu = 1$. We refer to (\ref{eq:soft_opt_pos}) with needed pre-computations as the \textit{soft optimization solution version}, simply denoted ``\textit{S.Opt.}''.

\subsection{Multi-AprilTag solutions}

All above presented solution versions of vehicle top tag pose estimation, i.e. \textit{Bas.}, \textit{H.Opt.}, \textit{S.Opt.}, can be directly applied to multi-AprilTag configurations and other kinds of vehicle top tag configurations, only if a proper set of tag control points can be provided. For example, consider the double-AprilTag configuration illustrated in Fig. \ref{fig:vehicle_top_tags}-top-right (which is preferred by us), one just needs to put the $n=8$ AprilTag control points $x_i$, $y_i$, $u_i$, $v_i$ ($i \in \{1 \cdots n\}$) 
\footnote{Pay attention that the two AprilTags share a common vehicle top tag coordinates system and their control points should be registered consistently.}
instead of the four specified in (\ref{eq:AprilTag_veh_top_xy}) and (\ref{eq:AprilTag_img_uv}) into the formalisms of the solution versions and does not need to do any change to the formalisms themselves.

\subsection{Multi-camera solutions}

The solution versions \textit{H.Opt.} and \textit{S.Opt.} can be easily extended to scenarios where multiple visual RSUs can perceive vehicle top tags. For example, if we replace the objective function in (\ref{eq:soft_opt_pos}) by the following objective function
\begin{equation}  \label{eq:soft_opt_multi_obj}
f(\mathbf{w T}) = \sum_{i=1}^n (\sum_{k=1}^m \| \begin{bmatrix} u_{i,k}^S \\ v_{i,k}^S \end{bmatrix} - \begin{bmatrix} u_{i,k} \\ v_{i,k} \end{bmatrix} \|_2^2 + \mu^2 |z_{w,i}^S - h|^2)
\end{equation}
then a multi-camera solution version is obtained. In (\ref{eq:soft_opt_multi_obj}), the subscript $k$ means image projections on the $k$-th visual RSU.

Intersections normally are indeed equipped with multiple visual RSUs instead of single one. It also makes sense that multiple visual RSUs tend to outperform a single visual RSU. On the other hand, we have found that using a single visual RSU (say the closest one to the vehicle) is already satisfactory. So in the spirit of \textit{Occam razor}, we exclude demonstration of multi-camera solutions in Section \ref{sec:simulation}.

\section{Simulation}  \label{sec:simulation}

Practical applications oriented simulation results are demonstrated. The simulated intersection is illustrated in Fig. \ref{fig:cam_four_views}. The lane width is $3.7$ meters and each road consists of four lanes in two directions. For the world coordinates system, its origin is located at the intersection center with its $x$-axis, $y$-axis facing along two perpendicular road directions and its $z$-axis facing upward. There are four visual RSUs symmetrically distributed at the four corners of the intersection respectively and each of them faces towards the intersection. For simulation, consider a generic one among them, which is supposed to be installed at a height of $8$ meters, looking downward by $40^\circ$. Further suppose all intrinsic and extrinsic parameters of the visual RSU are known. 

The autonomous public bus is set $6$ meters long, $2$ meters wide, and $3$ meters high. The AprilTag width is $1.6$ meters. As explained in Section \ref{sec:tag_pose_estimation}, the presented solutions are applicable to single-AprilTag and multi-AprilTag configurations. We prefer the double-AprilTag configuration such as illustrated in Fig. \ref{fig:vehicle_top_tags}-top-right, simulation results based on which are demonstrated below.

\subsection{Performance with moderate (low) resolution}  \label{sec:perf_low_res}

Suppose the visual RSU has a resolution of $960$-by-$720$, which is rather moderate or even low for visual RSUs. We intentional choose low-resolution visual RSUs in simulation to demonstrate practicality of the presented solutions, namely they can work with a large variety of visual RSUs --- They can even work with low-resolution visual RSUs and hence can naturally work with visual RSUs with higher resolutions --- We select a sector that can well cover the intersection quarter closest to the visual RSU, generate a large amount of ground-truth vehicle poses randomly in the sector and synthesize camera images according to the ground-truth vehicle poses respectively. For each synthesized image of the intersection scenario, we first use the method presented in Section \ref{sec:tag_detection} to detect bus top AprilTags and extract AprilTag control points which serve as common input to the various solution versions presented in Section \ref{sec:tag_pose_estimation}.

\begin{figure}[h!]
\begin{center}
\includegraphics[width=0.99\columnwidth]{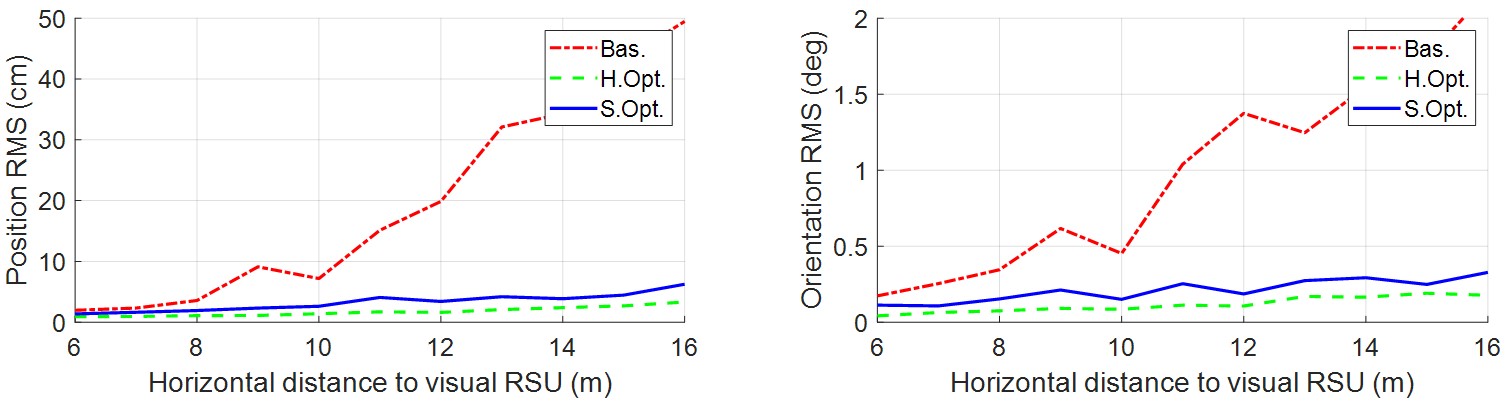}
\end{center}
\caption{Performance of vehicle top tag assisted localization by low-resolution visual RSU (no height disturbance)}
\label{fig:low_res_no_h_disturb}
\end{figure}

Performance statistics of the solution versions are obtained as follows: Horizontal integer meter distances of the bus center to the visual RSU are considered. Vehicle top tag assisted localization results associated with a horizontal integer meter distance, i.e. those associated with ground-truth vehicle poses that roughly have the integer meter distance to the visual RSU horizontally, are grouped to compute performance statistics associated with the integer meter distance. For example, for the solution versions (i.e. \textit{Bas.}, \textit{H.Opt.}, \textit{S.Opt.}), their position and orientation root mean square (RMS) errors associated with different horizontal integer meter distances are demonstrated in Fig. \ref{fig:low_res_no_h_disturb}.

For all of \textit{Bas.}, \textit{H.Opt.}, \textit{S.Opt.}, the closer the bus to the visual RSU is, the better the performance is. This is easily comprehensible: visual sensors have better perceptive ability towards closer objects than further objects. Performance of \textit{Bas.} deteriorates noticeably, yet it can still have sub-meter-level positioning performance at a distance of $16$ meters. \textit{H.Opt.} and \textit{S.Opt.} especially \textit{H.Opt.} have rather stable and ideal performance, thanks to the geometry constraint (\ref{eq:h_geometry_constraint}).

\begin{figure}[h!]
\begin{center}
\includegraphics[width=0.99\columnwidth]{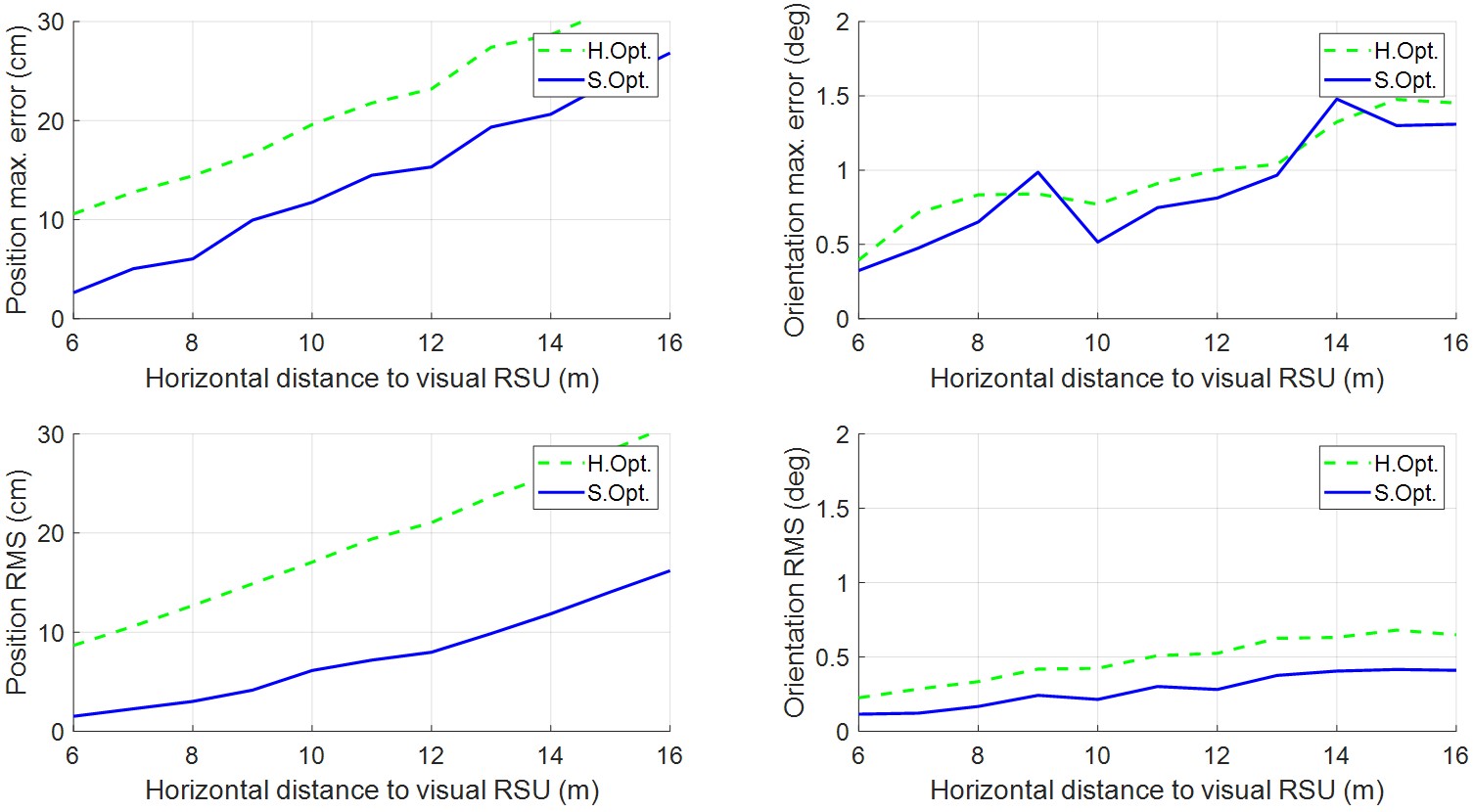}
\end{center}
\caption{Performance of vehicle top tag assisted localization by low-resolution visual RSU (with exaggerated height disturbance)}
\label{fig:low_res_with_h_disturb}
\end{figure}

Performance statistics demonstrated in Fig. \ref{fig:low_res_no_h_disturb} are obtained under the assumption that the bus has no height disturbance. In practice, the autonomous public bus indeed has moderate height disturbance especially when passing intersections
\footnote{Traffic rules require that vehicles move at low speeds ($\leq 30 \mbox{ km/h}$) when passing intersections. Moving at low speeds, vehicles especially buses which normally have good suspension systems tend to have rather stable height.}.
On the other hand, to test robustness of \textit{H.Opt.} and \textit{S.Opt.} against height disturbance, we intentionally exaggerate the height disturbance to $10$ centimetres. Relevant performance statistics are demonstrated in Fig. \ref{fig:low_res_with_h_disturb}.

\textit{S.Opt.} still has desirable performance when the vehicle is assumed to suffer from exaggerated height disturbance. Even its maximum positioning error at a distance of $16$ meters is below $30$ centimetres and its positioning RMS error at such distance is below $20$ centimetres. Besides, its orientation RMS error is below half a degree. The demonstrated raw performance of \textit{S.Opt.}, even under the influence of exaggerated bus height disturbance, is already satisfactory. Note that in practice such performance can be further augmented by proprioceptive sensors via data fusion.

\subsection{High resolution is power}  \label{sec:perf_high_res}

\begin{figure}[h!]
\begin{center}
\includegraphics[width=0.99\columnwidth]{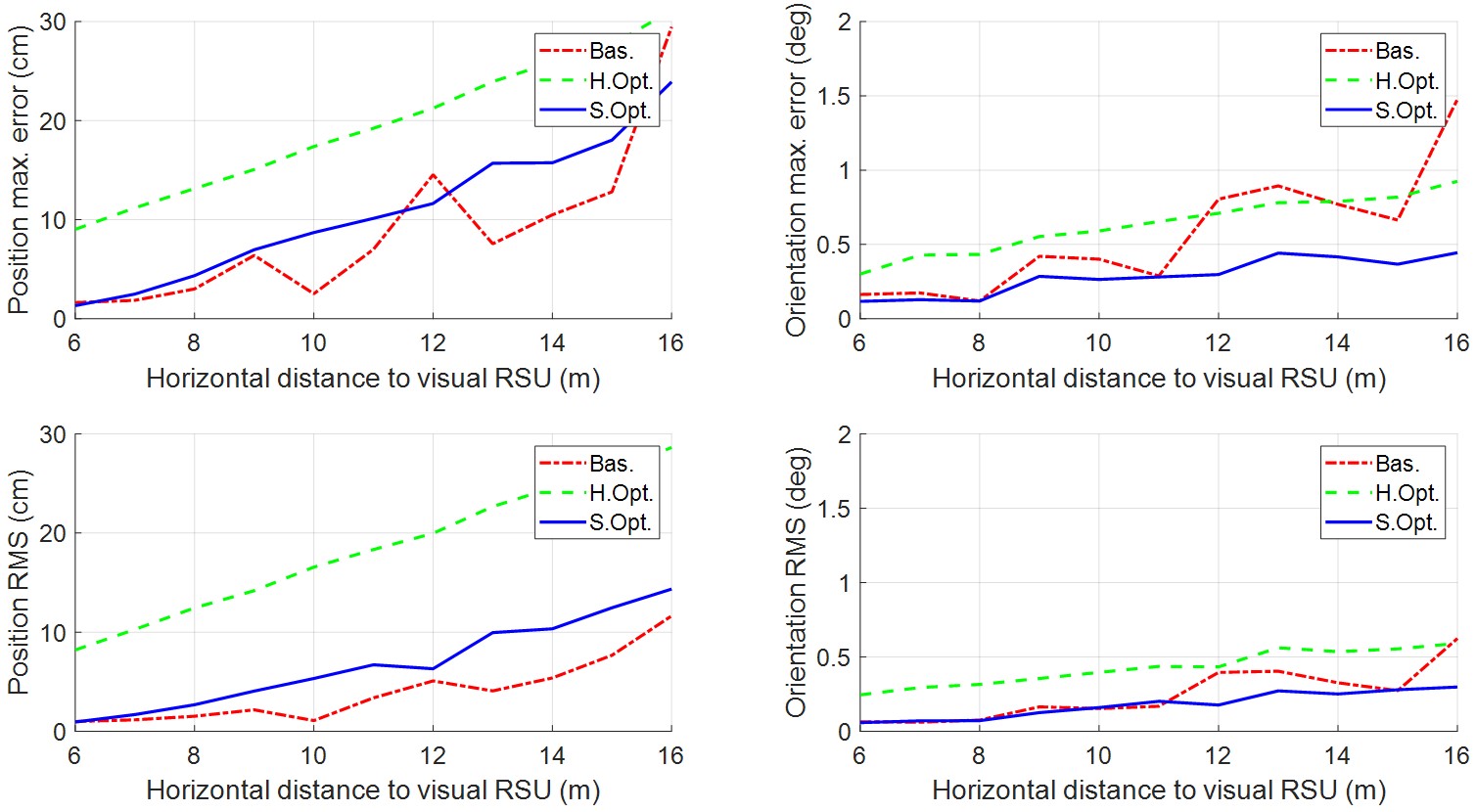}
\end{center}
\caption{Performance of vehicle top tag assisted localization by high-resolution visual RSU (with exaggerated height disturbance)}
\label{fig:high_res_with_h_disturb}
\end{figure}

Now suppose the visual RSU has a resolution of $3200$-by-$2400$ and perform similar simulation tests as presented in Sub-section \ref{sec:perf_low_res}. We still exaggerate the height disturbance to $10$ centimetres. Relevant performance statistics are demonstrated in Fig. \ref{fig:high_res_with_h_disturb}. High resolution of the visual RSU brings evident performance enhancement to all of \textit{Bas.}, \textit{H.Opt.}, \textit{S.Opt.} especially the first version.

\subsection{Discussion}

Providing various solutions of vehicle top tag pose estimation in Section \ref{sec:tag_pose_estimation} and demonstrating their performances in Section \ref{sec:simulation} are neither to scrutinize their nuances nor to advocate certain best one among them. Providing various solutions is to enable readers to have flexible choices according to their own practical engineering needs or preferences. In fact, \textit{Bas.}, \textit{H.Opt.}, \textit{S.Opt.} can all be incorporated holistically into the proposed methodology of vehicle top tag assisted localization and complement each other. For example, it seems that \textit{S.Opt.} is better than \textit{H.Opt.} as the former demonstrates more robustness against bus height disturbance than the latter. On the other hand, when the bus stops momentarily when turning right at the intersection
\footnote{Buses are big vehicles which are required so according to traffic rules.},
there is no height disturbance at all and \textit{H.Opt.} can be fairly utilized there. For another example, when visual RSUs are of high resolutions, \textit{Bas.} can be directly used whereas \textit{S.Opt.} can serve as ``double guarantee'' or ``cross validation''. After all, \textit{Bas.}, \textit{H.Opt.}, \textit{S.Opt.} involve only few analytical computations and sparse optimizations in iterative way, and they can be all executed at very little computational cost. 

Vehicle-road communication delay is moderate yet existing. However, this has almost negligible influence on vehicle top tag assisted localization, because \textit{back-projection} techniques (especially noting that \textit{Bas.}, \textit{H.Opt.}, \textit{S.Opt.} can provide estimates of the \textit{complete vehicle pose}) can be applied with proprioceptive sensors. Details are omitted here.

For administration reasons, we do not demonstrate real images but resort to demonstration via simulation, yet we believe this paper would be already clear enough for readers to capture key points and see merits of the proposed methodology of vehicle top tag assisted localization in valuable applications such as autonomous public buses.

\section{Conclusion}  \label{sec:conclusion}

We have proposed the methodology of \textit{vehicle top tag assisted vehicle-road cooperative localization} or for short \textit{vehicle top tag assisted localization}. We have reviewed a method of AprilTag detection, yet any other kind of vehicle top tags and relevant detection methods can be used for the proposed methodology. We have presented various solution versions of vehicle top tag pose estimation with extracted tag control points. Simulation results are provided to demonstrate effectiveness of the solution versions.

In our application context, we focus on handling difficult intersection scenarios for autonomous public buses, yet the proposed methodology of vehicle top tag assisted localization can be applied in much wider way. First, on the side of vehicles, the proposed methodology can be directly extended to practical applications concerning automation of other kinds of big vehicles. Second, on the side of roads, the proposed methodology can be directly extended to other road scenarios wherever visual RSUs are available, not limited only to intersection scenarios.

\section*{Appendix: vehicle top tag examples}

\begin{figure}[h!]
\begin{center}
\includegraphics[width=0.97\columnwidth]{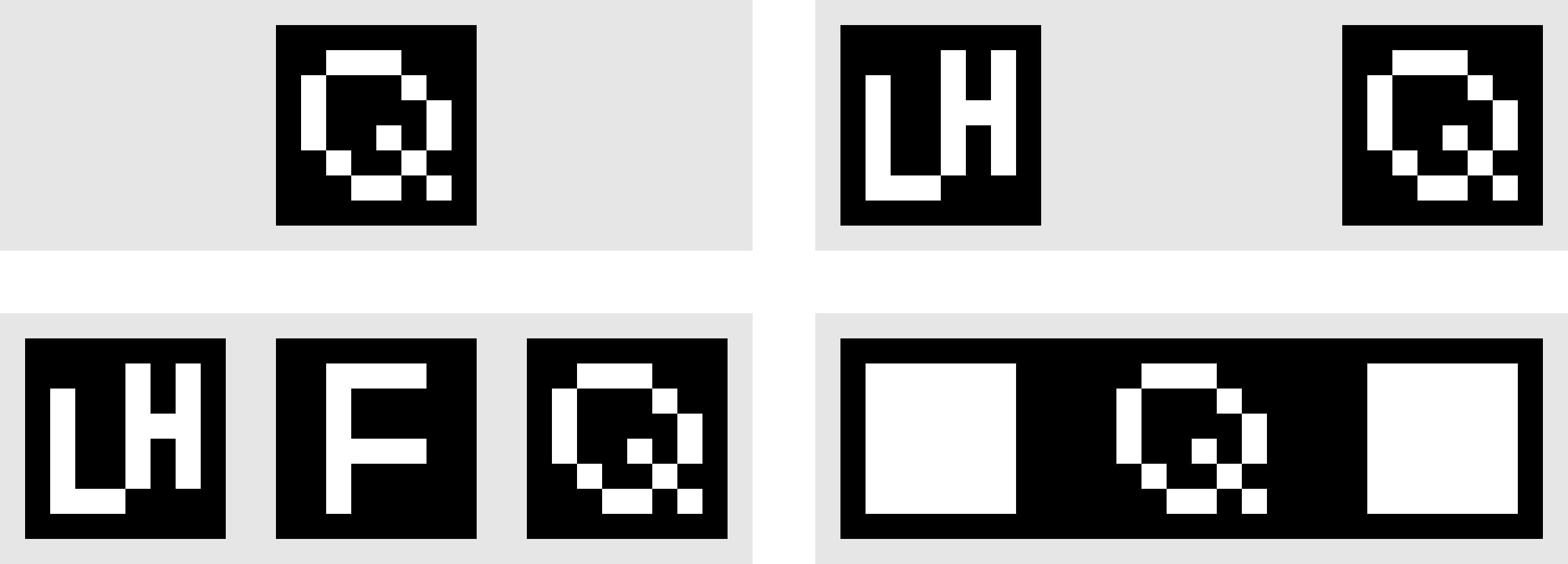}
\end{center}
\caption{Normal (bird-eye) views of vehicle top tag examples: (top-left) one tag installed at the bus top center; (top-right) two tags installed at the front and rear of the bus top respectively; (bottom-left) three tags installed at the front, center, and rear of the bus top respectively; (bottom-right) a long tag on the bus top}
\label{fig:vehicle_top_tags}
\end{figure}

Some examples of vehicle top tags are illustrated in Fig. \ref{fig:vehicle_top_tags}, where each sub-figure shows the normal (bird-eye) view of the autonomous public bus facing the left side.

\section*{Acknowledgement}

The company \textit{Qingfei.AI} deeply appreciates help and support from the government of Hangzhou.

\bibliographystyle{IEEEtran}
\bibliography{LI_Ref}

\end{document}